\documentclass[a4paper,fleqn]{cas-dc}
\usepackage[numbers,sort&compress]{natbib}
\usepackage{microtype}
\usepackage{url}
\usepackage{placeins}
\ExplSyntaxOn
\cs_set:Npn \__first_footerline: {\small\sffamily Working~manuscript~--~13~September~2026}
\ExplSyntaxOff
\newcommand{\StaticNMAE}{0.062028}
\newcommand{\GateNMAE}{0.062021}
\newcommand{\GBNMAE}{0.066130}
\newcommand{\GateAccuracy}{93.797912}
\newcommand{\StaticVsGB}{6.20}
\newcommand{\GateGain}{0.00000754}
\newcommand{\CILow}{-0.00001381}
\newcommand{\CIHigh}{0.00002845}

\shorttitle{Ensemble complexity in photovoltaic forecasting}
\shortauthors{Anonymous authors}
\begin{document}
\let\WriteBookmarks\relax
\title[mode=title]{Assessing Ensemble Complexity in Photovoltaic Power Forecasting}
\author[1]{Sun Ze}[style=chinese]
\author[1]{Zhou Liguo}[style=chinese]
\author[1]{Xu Yuqing}[style=chinese]
\author[1]{Yu Lei}[style=chinese]
\author[1]{Jiang Mingming}[style=chinese]
\affiliation[1]{organization={School of Computer Science and Technology, Huaibei Normal University},country={China}}
\begin{abstract}
An ensemble can improve photovoltaic forecasts while adding components that contribute little or increase computation. We assess these effects through matched comparisons and ablations of a fixed heterogeneous predictor bank. Hourly experiments use GEFCom2014 and three additional public datasets, with chronological partitions and three seeds. Under retrospective ERA5 assistance, static fusion reduces scaled mean absolute error against matched boosting by 1.11\%, 4.41\%, and 1.63\% on PVDAQ, OPSD, and Ausgrid; only OPSD remains supported after multiple-comparison correction. Weather gating offers no consistent incremental benefit. Exploratory member removals show group-level dependence alongside individual redundancy. A separate, previously inspected fifteen-minute case replaces one neural member with a tree predictor: normalized error falls by 1.72\%, but measured inference is slower. These findings support component-wise evaluation with explicit limits on weather availability and test-set reuse.
\end{abstract}
\maketitle

\section{Introduction}
Photovoltaic (PV) power forecasts over the next few hours support the scheduling of generation, storage, and reserve capacity. At this horizon, the daily solar cycle is predictable, whereas passing clouds and site-specific operating conditions can change output rapidly. Forecast errors therefore depend on the prediction horizon, local measurements, and the weather information available when the forecast is issued \citep{antonanzas2016,pvreview2022}. These differences also complicate comparisons: target-period realized weather provides information that a model restricted to past observations would not have.

Forecast combination is an established response to uncertainty about the best predictor \citep{wang2023}. A bank can contain recurrent networks, attention-based models, temporal mixers, and regression predictors, followed by learned weights, site-specific adjustments, and weather-dependent corrections. Each component adds something to the implementation; it need not add useful predictive information. The practical question is whether the resulting error reduction justifies the added fitting and inference requirements.

Attribution is particularly difficult when an ensemble is compared with a single predictor. A difference can arise from combining forecasts, from extra weather inputs, or from labels used to fit the weights but withheld from the control. An equal epoch count per neural member also does not equalize the total cost of training a bank and training one model. Component ablations and explicit information and fitting-period controls are therefore needed to interpret an aggregate score \citep{hewamalage2023}. The incremental analysis of NWP-to-power processing by \citet{mayer2025complexity} motivates a similar question here, although our ensemble members are power predictors rather than weather forecasts.

We study three questions. First, does learned fusion improve on equal averaging and a regression control with matched information and fitting-label access? Second, do site dependence, global shrinkage, and weather gating contribute beyond simpler combinations, and how sensitive is the current combination to its members? Third, what accuracy and inference-cost change follows the replacement of one member in a separately evaluated small ensemble? These questions concern the value of specific components; model count alone is not treated as a methodological contribution.

The primary implementation, denoted V9 in the source project, learns station--horizon convex weights, shrinks them toward a global estimate, and applies a bounded weather-regime adjustment with a chronological confirmation rule. Its archived private score of approximately 91.8\% is audited separately. Public experiments train a fourteen-member hourly adaptation on GEFCom2014 \citep{hong2016,dumas2022} and three further sources. The latter experiments use matched regression controls and explicitly retrospective ERA5 weather. We retain failed gate confirmations and perform exploratory fixed-weight removals without selecting a new model from their test results.

A secondary fifteen-minute case compares two four-slot combinations on identical private targets, retaining three neural checkpoints and replacing the fourth member with a tree predictor. It adds a measured inference-cost comparison to the accuracy analysis, but uses an already inspected historical period and is not an independent public validation. The contributions are measured accuracy gains within explicitly controlled comparisons, component-level evidence explaining their limits, and an inference-cost assessment. Member count is a design variable in this assessment; the different implementations do not establish a controlled reduction from 82 to 14 to four models.

\section{Related works}
\subsection{PV forecasting and weather-conditioned models}
The survey of \citet{antonanzas2016} provides a foundation for distinguishing PV forecasting tasks by their inputs and horizons. \citet{pvreview2022} review data-driven PV forecasting procedures, while the more recent deep-learning review of \citet{deepreview2024} identifies inconsistent datasets, forecasting horizons, and evaluation metrics as obstacles to comparing published models. These observations motivate evaluating all constituents and ensemble variants on a shared target population.

Weather classification has already been investigated as a forecasting mechanism. \citet{weatherensemble2020} combine random forests, support vector regression, and deep belief networks, with cloud-based classification and separate ensembles for different subsets. Their study uses 21 German PV facilities at three-hour resolution and a shuffled train--test split. Weather-classification-based MARS forecasting provides another example of this research direction \citep{weathermars2019}. Consequently, weather-conditioned aggregation itself should not be presented as a new idea.

The implementation studied here uses a different parameterization: it keeps its predictor bank fixed, first learns local and global convex weights, and then learns bounded regime-dependent offsets at the model-family level. This structure avoids training a separate copy of every model for every weather class. Whether this smaller adjustment is useful remains an empirical question. Published errors from weather-classification studies are not used as numerical baselines here because their datasets, resolutions, splits, and meteorological information differ.

Forecast meteorology also differs from realized meteorology. Work combining ensemble numerical weather prediction (NWP) with physical model chains explicitly considers forecast-weather uncertainty \citep{weatherchain2023}. An archived dataset from the ECMWF Ensemble Prediction System offers a further example of data developed specifically for probabilistic solar power forecasting \citep{nwpdataset2022}. It is a relevant source for future evaluation, rather than an additional dataset tested in this study. This distinction matters for the private V9 experiment because its weather context is obtained from a historical weather cache at target timestamps. In contrast, the GEFCom2014 solar variables are supplied as NWP predictors, although the processed mirror used here does not retain forecast-issuance timestamps. Both limitations are disclosed in the experimental design.

The choice of processing steps also deserves attention. \citet{mayer2025complexity} compare sixteen workflows for converting ensemble NWP into deterministic PV forecasts at five Hungarian plants. Their analysis shows that some intermediate processing stages add little once the final power forecast is bias-corrected. This finding motivates evaluating the incremental contribution of each stage, rather than attributing the final error to every component in a pipeline. Their ensemble consists of weather-forecast members, whereas the present study combines separately trained power predictors. The datasets, information sets, and error normalization also differ, so their reported scores are not numerical baselines for this study.

\subsection{Neural and regression forecasting models}
Long short-term memory networks model sequential dependencies through gated recurrent states \citep{hochreiter1997}. Temporal Fusion Transformers provide an interpretable multi-horizon forecasting architecture \citep{tft2021}. PatchTST introduces patch-based time-series representations \citep{nie2023}, while TiDE uses a dense encoder--decoder with covariates \citep{das2023}. TSMixer and TimeMixer study mixing operations along feature, temporal, or sampling-scale dimensions \citep{chen2023,timemixer2024}. Mamba introduces input-dependent selective state-space dynamics as an alternative sequence-modeling mechanism \citep{mamba2023}; its general sequence results do not establish performance on this PV task. These methods motivate architectural diversity, rather than a presumption that one architecture dominates PV forecasting.

The findings of \citet{zeng2023} also motivate including simple models in time-series comparisons. Accordingly, the public experiment includes ridge regression and gradient boosting \citep{friedman2001}, together with persistence-based baselines. The neural benchmarks are the original project's implementations of nine architecture families. Several are explicitly simplified or task-adapted: for example, the repository uses a pure-PyTorch Mamba-style block. The benchmark therefore compares this implementation family and does not establish a ranking of the authors' official implementations of all named architectures.

\subsection{Forecast combinations and evaluation}
Forecast combination spans simple averages, learned weighting, and conditional combinations \citep{wang2023}. A station-specific weight can capture local differences, while a horizon-specific weight can reflect the changing usefulness of persistence and meteorological predictors. Shrinkage toward global weights is a practical way to reduce the sensitivity of local estimates when few fitting windows are available.

Input-dependent allocation among predictors also has a long history. Adaptive mixtures of local experts learn to assign subsets of examples to expert networks \citep{jacobs1991}, and hierarchical mixtures of experts model both mixture coefficients and component responses within a learned hierarchy \citep{jordan1994}. V9 belongs to the broader family of conditional combinations, but the evaluated implementation fits a bounded adjustment to a previously trained predictor bank. It does not jointly train a hierarchical mixture model or use the EM procedure of \citet{jordan1994}. Its usefulness must therefore be assessed through the specific weight parameterization and confirmation protocol, rather than through a claim that conditional expert weighting is new.

Evaluation must separate predictor fitting, hyperparameter selection, ensemble fitting, and final scoring \citep{hewamalage2023}. GEFCom2014 provides a widely used public setting for renewable-energy forecasting \citep{hong2016}; it has also supported probabilistic generative-model studies \citep{dumas2022}. The present task is a deterministic, rolling four-hour adaptation of the solar track. Competition quantile scores and published probabilistic results are therefore contextual references, rather than directly comparable scores.

\section{Method}
\subsection{Task and information set}
Let $p_{s,t}$ denote the output of site $s$ and let $c_s>0$ be its normalization scale. The target is $y_{s,t}=p_{s,t}/c_s$. Given $L$ historical observations, calendar variables, and an eligible meteorological input $\boldsymbol{x}$, predictor $m$ produces
\begin{equation}
 \widehat y^{(m)}_{s,t+h}=f_m(\boldsymbol{y}_{s,t-L+1:t},\boldsymbol{x},s,h),
 \qquad h=1,\ldots,H.
\end{equation}
The private task uses five-minute data and $H=48$; the public adaptation uses hourly data and $L=24$, $H=4$. Both forecast the next four hours. The separate native fifteen-minute replacement case is specified in Section~\ref{sec:native}; it is not a further compression of this bank. In a deployed system, $\boldsymbol{x}$ must belong to the information set available at origin $t$. The private archived result is treated as a retrospective weather-assisted result until this requirement is established for its weather feed.

\subsection{Frozen heterogeneous predictor bank}
The archived V9 bank contains 72 trained predictors and 10 statistical baselines. Its learned predictors span nine neural architecture families and a station-wise ridge component; checkpoints differ in training history and weather usage. The baseline groups include persistence, daily persistence, climatology, state-conditioned climatology, and several historical-analog configurations. V9 changes the combination weights while leaving these predictors frozen.

The public experiment retrains one implementation from each of the nine neural families per random seed and adds five classical predictors. This gives $M=14$ constituents per run. Public sites use new embeddings and newly fitted models; private weights and checkpoints are not transferred to the public test set. Reducing the bank is an explicit experimental adaptation, so the public experiment validates the combination mechanism at this scale rather than reproducing the private 82-constituent system.

\subsection{Station--horizon weights and shrinkage}
Predictor errors can vary by site and lead time. We represent this variation with nonnegative weights, parameterized by logits and normalized across predictors:
\begin{equation}
 w_{s,m,h}=\frac{\exp(a_{s,m,h})}{\sum_{j=1}^{M}\exp(a_{s,j,h})}.
\end{equation}
The weights minimize mean absolute error on the ensemble-fitting partition. Local estimates can be unstable when a site contributes few windows, so a separate global vector $w^{(g)}_{m,h}$ is fitted using all sites. Each local estimate is shrunk toward this shared vector:
\begin{align}
 \bar w_{s,m,h}&=\alpha_s w_{s,m,h}+(1-\alpha_s)w^{(g)}_{m,h},\\
 \alpha_s&=\frac{n_s}{n_s+\kappa},
\end{align}
where $n_s$ counts fitting windows and $\kappa$ is selected on a later chronological partition. Convexity is preserved because both component weight vectors sum to one. The static ensemble is $\widehat y^{(0)}_{s,t+h}=\sum_m \bar w_{s,m,h}\widehat y^{(m)}_{s,t+h}$.

\subsection{Weather regimes}
The selected private configuration, \texttt{physical\_weather3}, uses irradiance index $k$, cloud cover $q$ in percent, precipitation $r$, and the absolute one-hour change $d_k$. The three regimes are assigned by
\begin{equation}
 z=\begin{cases}
 0,& k\geq0.75,\ q\leq50,\ r<0.05,\ d_k<0.15,\\
 2,& k<0.35\ \text{or}\ q\geq70\ \text{or}\ r\geq0.05,\\
 1,& \text{otherwise}.
 \end{cases}
\end{equation}
The adverse regime has precedence in the source implementation. The private context uses the cached clear-sky index and target-time weather. These regime names describe deterministic rules; they are not independently verified weather labels.

GEFCom2014 does not provide site coordinates in the mirror used here. The public adaptation therefore replaces the physical clear-sky index with an empirical radiation-envelope index. For each site, calendar month, and hour, the 95th percentile of training-period forecast surface irradiance defines $e_{s,m,h}$. The index is
\begin{equation}
 k^{(e)}=\operatorname{clip}\!\left(\frac{\mathrm{SSRD}}{\max(e_{s,m,h},20)},0,2\right).
\end{equation}
This quantity is not a physical clear-sky index. Its denominator uses only the base-training period. The supplied cloud fraction is multiplied by 100 and hourly precipitation in metres by 1,000 before applying the existing thresholds. An origin-regime ablation repeats the last historical regime over all four horizons while keeping the base forecasts unchanged.

\subsection{Bounded family-level weather gate}
Static weights summarize performance over the fitting period, but they cannot respond to a change in weather regime. The gate adds this response without retraining the predictors. Let $g(m)$ map each predictor to a family. Its weighted contribution and weight mass are
\begin{align}
 C_{s,t,g,h}&=\sum_{m:g(m)=g}\bar w_{s,m,h}\widehat y^{(m)}_{s,t+h},\\
 B_{s,g,h}&=\sum_{m:g(m)=g}\bar w_{s,m,h}.
\end{align}
For regime $z$, horizon $h$, and family $g$, an unconstrained parameter $b_{z,h,g}$ is bounded and centered:
\begin{align}
 u_{z,h,g}&=0.75\tanh(b_{z,h,g}),\\
 \delta_{z,h,g}&=u_{z,h,g}-G^{-1}\sum_{j=1}^{G}u_{z,h,j}.
\end{align}
The gated forecast becomes
\begin{equation}
 \widehat y^{(z)}_{s,t+h}=\operatorname{clip}\!\left(
 \frac{\sum_g e^{\delta_{z,h,g}}C_{s,t,g,h}}
 {\max(\sum_g e^{\delta_{z,h,g}}B_{s,g,h},10^{-7})},0,1.25\right).
\end{equation}
This renormalized form preserves nonnegative constituent weights. A zero gate exactly recovers the static combination for forecasts within the clipping interval; this identity is checked against the saved implementation. The source also clips exponent arguments to $[-2,2]$ during application. The gate is fitted by minimizing
\begin{equation}
 \mathcal L=\frac{1}{|\Omega|}\sum_{(s,t,h)\in\Omega}|\widehat y^{(z)}_{s,t+h}-y_{s,t+h}|
 +\lambda\operatorname{mean}(\delta^2),
\end{equation}
where $\Omega$ contains valid fitting targets. The archived gate has $3\times48\times20=2{,}880$ stored coefficients. The public adaptation has $3\times4\times14=168$ because each constituent is its own family in a run.

\begin{figure*}[t]
\centering
\includegraphics[width=\textwidth]{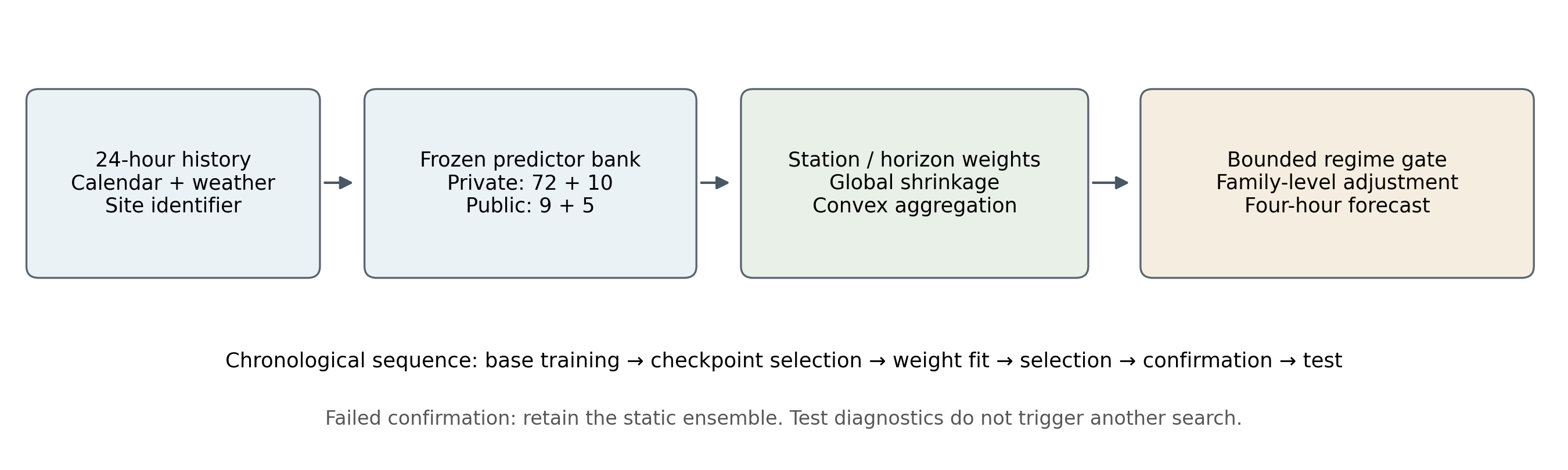}
\caption{Forecast combination and evaluation. Base predictors are frozen before weight learning. Weather availability must be established at the forecast origin. The schematic was drawn with reproducible Python code prepared with OpenAI Codex assistance (OpenAI, September 2026).}
\label{fig:method}
\end{figure*}

\subsection{Selection and confirmation}
The gate is accepted only if its improvement persists beyond the period used to select its regularization. To make this check possible, the public pipeline separates base training, checkpoint selection, weight fitting, parameter selection, confirmation, and testing in chronological order. The shrinkage grid is $\{250,500,1000,2000,4000\}$ and the gate regularization grid is $\{0.01,0.1,1\}$. The archived V9 compared three regime definitions; the public experiments fix its retained three-regime definition before testing.

The confirmation period is divided into three consecutive blocks. A candidate passes only if it improves MAE in all three blocks and the mean improvement is at least $10^{-6}$. Once choices are frozen, weights are refitted on the combined fitting, selection, and confirmation periods. The primary safeguard falls back to static weights if confirmation fails. For transparency, the experimental tables also report the rejected candidate's test error as a diagnostic. These diagnostic errors do not trigger additional tuning.

\subsection{Fixed-weight member-removal diagnostic}
To examine dependence on the fitted bank, we remove a member or group $R$ from the static combination and renormalize the remaining station--horizon weights:
\begin{equation}
 \widehat y^{(-R)}_{s,t+h}=
 \frac{\sum_{m\notin R}\bar w_{s,m,h}\widehat y^{(m)}_{s,t+h}}
 {\sum_{m\notin R}\bar w_{s,m,h}}.
\end{equation}
The retained weight mass is positive in every evaluated case. We report $\Delta_R=\mathrm{sMAE}(\widehat y^{(-R)})-\mathrm{sMAE}(\widehat y^{(0)})$, so positive values mean that removal worsens the current combination. Base predictions and remaining relative weights are fixed; neither refitting nor gating is performed.

The diagnostic covers each of the fourteen members, all nine neural members together, and all five classical members together. The full static reference plus sixteen removals produce seventeen scenarios for each dataset and seed. This analysis was specified after the existing test results had been inspected and is explicitly exploratory. It measures dependence on frozen predictions, not the best performance of a smaller retrained bank, an optimal number of models, or independent evidence for test-driven pruning.


\section{Experiments}
\subsection{Evidence scope}
The evaluation has three distinct settings. The private five-minute V9 archive verifies implementation identity and the reported score, without retraining its full bank. The hourly public study tests the retained combination mechanism on GEFCom2014 and, with stronger information and label controls, PVDAQ, OPSD, and Ausgrid. The added fixed-weight removals use those already inspected public tests as exploratory diagnostics. Finally, a separate native fifteen-minute private case examines a matched member replacement and its measured inference cost. Absolute scores and member counts are not ranked across these settings because targets, inputs, fitting histories, and normalization differ.


\subsection{Archived private case study}
The retained V9 configuration is identified by the saved summary, the weights array, and its matching training log. Its core source snapshot has SHA-256 prefix \texttt{ba5c8010b517178e}. The weight array has dimensions $(28,82,48)$ and the gate array $(3,48,20)$. Independent checks verify finite nonnegative weights, normalization, zero-gate identity, and the arithmetic relating nMAE to the reported percentage.

The archived snapshot is frozen at 25 August 2026 and contains 441,844 training windows, 184,980 validation windows, and 114,262 test windows. All 48 target timestamps of each evaluated window lie between 06:00 and 20:00. The test contains 5,484,576 window--horizon targets, including repeated timestamps arising from overlapping forecasting origins. These are not 5.48 million independent observations.

\begin{table}[t]
\centering
\caption{Archived private results on the common cleaned target population. No new private-data rerun is claimed.}
\label{tab:private}
\small
\begin{tabular}{lrr}
\toprule
Configuration & nMAE & Score (\%) \\
\midrule
Frozen reference & 0.08181196 & 91.818804 \\
V9 static control & 0.08178049 & 91.821951 \\
V9 gate (retained) & 0.08173462 & 91.826538 \\
\bottomrule
\end{tabular}
\end{table}


The V9 gain over its recomputed static control is 0.004587 percentage points in the reported score, corresponding to approximately 0.0561\% relative nMAE reduction.

The private results in Table~\ref{tab:private} are archived evidence. Neither all 72 base models nor their full private test predictions were regenerated for this manuscript. Earlier raw-data and cleaned-data scores use different target sets and are excluded from an improvement calculation. Moreover, several historical versions inspected the same private test period. The private test is consequently not a globally untouched benchmark across the entire development history. Its normalization provenance and target-time historical-weather availability remain additional limits on operational interpretation.

\subsection{GEFCom2014 dataset and forecasting protocol}
The public experiment uses \texttt{solar\_new.csv} from the published research repository accompanying \citet{dumas2022}. The downloaded commit and file hashes are stored with the experiment. The file contains 59,040 complete hourly records across three solar sites, with 19,680 records per site from 2 April 2012 01:00 to 1 July 2014 00:00. Duplicate timestamps and missing numeric values are absent in this file. POWER is already normalized by capacity; the observed value slightly above one at one site is retained.

The source preprocessing differences accumulated radiation and precipitation fields and shifts the complete site series by ten rows. The present experiment uses the supplied processed timestamps and does not apply another time-zone shift. Calendar-hour filtering therefore refers to the mirror's clock. The 12 meteorological channels include cloud water and ice, pressure, humidity, cloud cover, wind components, temperature, surface and top-of-atmosphere radiation, and precipitation.

Weather scaling uses training-only means and standard deviations; standardized channels are clipped to $[-10,10]$. Neither public power nor its evaluation denominator is rescaled using test observations. A valid forecasting window contains 24 past hours and four subsequent target hours; only the targets must lie within the clock interval $[06{:}00,20{:}00)$. Targets crossing a partition boundary are excluded, while earlier observations may legitimately serve as historical inputs for a later forecast.

\begin{table*}[t]
\centering
\caption{Chronological public partitions. Complete forecast targets must remain inside a partition.}
\label{tab:splits}
\small
\begin{tabular}{llr}
\toprule
Purpose & Period & Windows \\
\midrule
Base training & 2 Apr 2012--31 May 2013 & 13,992 \\
Checkpoint selection & June 2013 & 990 \\
Weight fitting & July--September 2013 & 3,036 \\
Hyperparameter selection & October--November 2013 & 2,013 \\
Confirmation & December 2013 & 1,023 \\
Final test & January--June 2014 & 5,973 \\
\bottomrule
\end{tabular}
\end{table*}


The final test covers January--June 2014 and contains 5,973 windows, or 23,892 window--horizon targets. Each site contributes 1,991 windows. All methods use the same targets. This is a within-site temporal evaluation across three public sites, not a held-out-site experiment. The public NWP variables are treated as supplied forecast covariates. Because their issue timestamps are absent, the experiment does not independently certify the latest forecast vintage available at every origin.

\subsection{Baselines and implementation}
The nine neural constituents are LSTM, TimeMixer-style, Transformer-style, Mamba-style, PatchTST-style, TiDE-style, TSMixer-style, ModernTCN-style, and TFT-style implementations from the project. Each uses a hidden dimension of 64, dropout 0.1, a 24-hour history, and the available weather and calendar inputs. PatchTST-style inputs use an eight-hour patch and a four-hour stride; TimeMixer scales are $1,2,4,8$. The original implementations are preserved in the reproducibility package, including their defaults and task-specific modifications.

For each seed, base models are trained for 12 epochs with AdamW, learning rate 0.002, weight decay $10^{-4}$, batch size 256, and gradient-norm clipping at 1. Each epoch visits the available training windows in a randomly permuted order; the configured 16,000-window cap exceeds the 13,992 available windows. Checkpoints are selected by the independent early-stopping period rather than by test error. Seeds 2026, 2027, and 2028 define the three runs, with fixed family-specific offsets.

The classical constituents are last-value persistence, previous-day same-hour persistence, training-only hourly median climatology, ridge regression with penalty 10, and histogram gradient boosting. Ridge and boosting use the historical power vector, future forecast covariates and calendar features, and site indicators. Four boosting regressors use absolute-error loss, 150 iterations, at most 15 leaves, $L_2$ regularization 1, and no internal early-stopping split. These deterministic classical predictions are shared across the three neural seeds. Equal per-member epochs limit neural fitting but do not match total bank training, hyperparameter search, or hardware costs to those of a single regression control. Base, static-ensemble, and gated-ensemble forecasts are clipped to $[0,1.25]$ consistently where models can exceed this range.

Weight optimization reuses the project's Adam-based simplex solver for 800 steps at learning rate 0.08. The gate uses the original 600-step routine, learning rate 0.03, and batches of up to 4,096 windows. Training runs locally with Python 3.11.16, PyTorch 2.6.0+cu124, and an NVIDIA GeForce RTX 3060 Laptop GPU.

Experiment and figure scripts were prepared with OpenAI Codex assistance (OpenAI, September 2026) and preserved with software and data hashes. Data plots were generated from saved predictions; independent routines reconstructed input windows, reported errors, and ensemble outputs.

\subsection{Metrics and uncertainty}
For the common target population $\Omega$, the main metrics are
\begin{align}
 \mathrm{nMAE}&=|\Omega|^{-1}\sum_{\Omega}|\widehat y-y|,\\
 \mathrm{nRMSE}&=\sqrt{|\Omega|^{-1}\sum_{\Omega}(\widehat y-y)^2}.
\end{align}
The project's score is $A=100(1-\mathrm{nMAE})$. It is a transformed regression error, not a classification accuracy or the percentage of forecasts within a tolerance. An additional active-output metric evaluates targets with $y>0.02$; this is a retrospective reporting subset, not a weather classifier or a rule for selecting the main test set.

Tables report means and sample standard deviations over three neural seeds. For the gate--static comparison, losses are averaged over seeds and aggregated by forecast-origin calendar day. A paired bootstrap resamples 181 complete day blocks 2,000 times. Each block retains all sites and overlapping horizons, reducing the false precision of pointwise resampling. The interval is conditional on this test period and the fitted models; it does not fully model dependence across consecutive days or uncertainty from alternative temporal splits.

\subsection{Original GEFCom2014 comparison results}
\begin{table*}[t]
\centering
\caption{Public benchmark on the same 5,973 test windows. Mean and sample standard deviation over three neural seeds.}
\label{tab:public}
\small
\begin{tabular}{lcccc}
\toprule
Method & nMAE & nRMSE & Active-output nMAE & Score (\%) \\
\midrule
Persistence & $0.232861 \pm 0.000000$ & $0.306324 \pm 0.000000$ & 0.242035 & 76.7139 \\
Daily persistence & $0.125993 \pm 0.000000$ & $0.202769 \pm 0.000000$ & 0.140266 & 87.4007 \\
Climatology & $0.153293 \pm 0.000000$ & $0.219130 \pm 0.000000$ & 0.168680 & 84.6707 \\
Ridge regression & $0.083104 \pm 0.000000$ & $0.116123 \pm 0.000000$ & 0.090445 & 91.6896 \\
Gradient boosting & $0.066130 \pm 0.000000$ & $0.106951 \pm 0.000000$ & 0.073832 & 93.3870 \\
LSTM & $0.071006 \pm 0.000537$ & $0.112797 \pm 0.000462$ & 0.078321 & 92.8994 \\
TimeMixer-style & $0.069744 \pm 0.000752$ & $0.111522 \pm 0.000860$ & 0.077052 & 93.0256 \\
Transformer-style & $0.070440 \pm 0.002103$ & $0.112701 \pm 0.003440$ & 0.078129 & 92.9560 \\
Mamba-style & $0.067656 \pm 0.000312$ & $0.109494 \pm 0.000868$ & 0.074763 & 93.2344 \\
PatchTST-style & $0.072165 \pm 0.003470$ & $0.111042 \pm 0.004843$ & 0.079617 & 92.7835 \\
TiDE-style & $0.067469 \pm 0.001019$ & $0.108096 \pm 0.002864$ & 0.074778 & 93.2531 \\
TSMixer-style & $0.067534 \pm 0.001537$ & $0.108203 \pm 0.002222$ & 0.074954 & 93.2466 \\
ModernTCN-style & $0.068819 \pm 0.001170$ & $0.109171 \pm 0.001475$ & 0.076030 & 93.1181 \\
TFT-style & $0.069498 \pm 0.000860$ & $0.112185 \pm 0.000809$ & 0.077275 & 93.0502 \\
Uniform ensemble & $0.069833 \pm 0.000172$ & $0.104328 \pm 0.000256$ & 0.075985 & 93.0167 \\
Static ensemble & $0.062028 \pm 0.000262$ & $0.102127 \pm 0.000600$ & 0.068987 & 93.7972 \\
V9 gate (hourly adaptation) & $0.062021 \pm 0.000263$ & $0.102134 \pm 0.000597$ & 0.068984 & 93.7979 \\
Origin-regime gate & $0.062012 \pm 0.000264$ & $0.102112 \pm 0.000603$ & 0.068969 & 93.7988 \\
Confirmation-selected rule & $0.062028 \pm 0.000262$ & $0.102127 \pm 0.000600$ & 0.068987 & 93.7972 \\
\bottomrule
\end{tabular}
\par\vspace{3pt}\footnotesize The style suffix denotes a repository implementation. Classical baselines are deterministic and shared across runs. The candidate gate and origin-regime variant are diagnostics; the confirmation-selected rule uses static weights in all runs.
\end{table*}


The static ensemble achieves nMAE \StaticNMAE{}, compared with \GBNMAE{} for gradient boosting, a relative reduction of \StaticVsGB{}\%. The nine neural constituents range from approximately 0.06747 to 0.07217 in mean nMAE. Thus, no individual neural family beats gradient boosting under the fixed budget, while the learned heterogeneous combination improves over both. Equal weighting is weaker than learned static weighting, indicating that diversity alone is insufficient when constituent error levels differ substantially. This is the original benchmark comparison: its regression controls omit historical weather channels and do not receive the later labels used for ensemble fitting. These differences limit architectural attribution of the reported improvement. The added experiments below use complete-feature, matched-label controls.

\begin{figure*}[t]
\centering
\includegraphics[width=\textwidth]{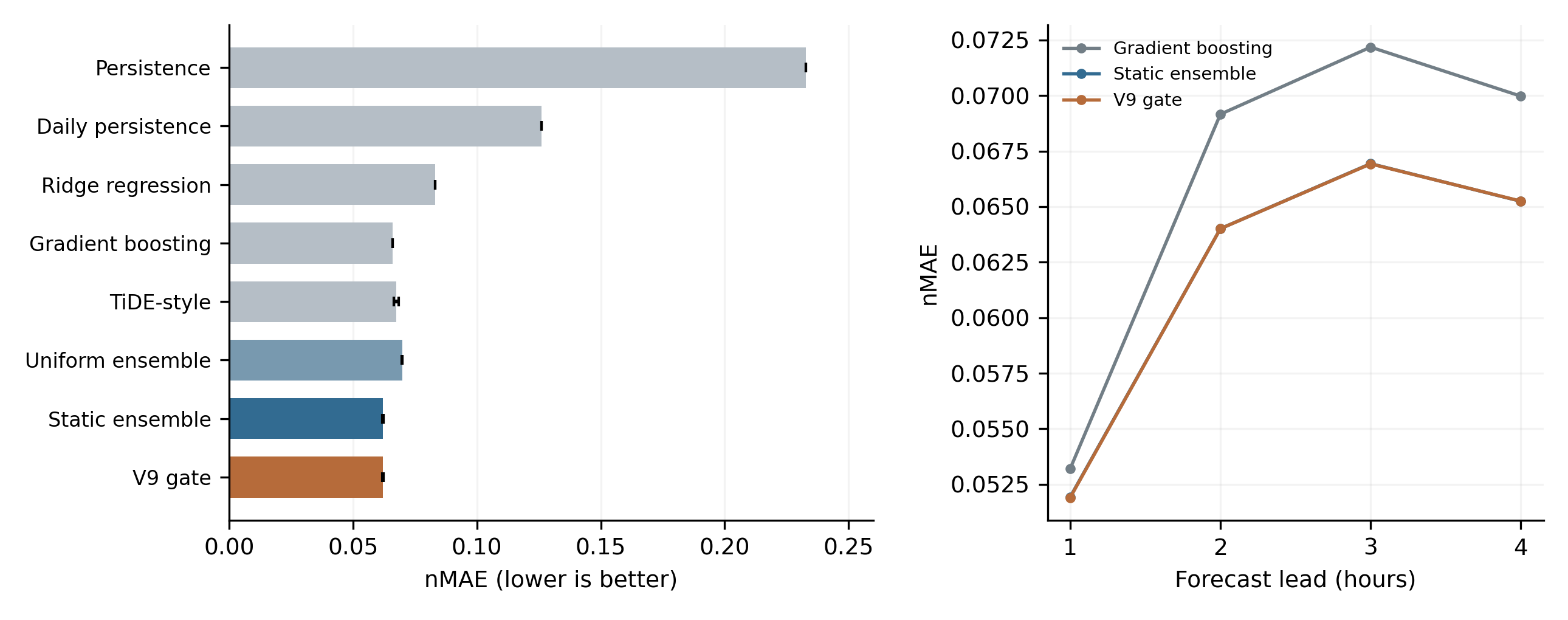}
\caption{Public-data comparison. Left: selected methods' mean nMAE; error bars show the sample standard deviation across three neural seeds. Deterministic baselines have no seed variation. Right: per-horizon nMAE for the strongest classical constituent and ensemble variants. All curves use identical test windows.}
\label{fig:public}
\end{figure*}

The target-weather gate has mean nMAE \GateNMAE{} and score \GateAccuracy{}\%. Its improvement over static weights is only \GateGain{} in normalized MAE, or approximately 0.000754 percentage points in the score. The 95\% day-block bootstrap interval is $[\CILow{},\CIHigh{}]$, which includes zero. In addition, the gate fails confirmation for each of the three training seeds. The predeclared confirmation-selected rule consequently uses the static ensemble in every run.

\begin{table*}[t]
\centering
\caption{Gate selection and final test diagnostics. Positive test gain means lower gated nMAE.}
\label{tab:gates}
\small
\begin{tabular}{rrrrrrl}
\toprule
Seed & $\kappa$ & $\lambda$ & Static nMAE & Gated nMAE & Test gain & Confirmation \\
\midrule
2026 & 1000 & 0.1 & 0.06174977 & 0.06174567 & +0.00000410 & Reject \\
2027 & 4000 & 1.0 & 0.06226879 & 0.06226920 & -0.00000041 & Reject \\
2028 & 4000 & 0.01 & 0.06206670 & 0.06204777 & +0.00001893 & Reject \\
\bottomrule
\end{tabular}
\end{table*}


The origin-regime ablation has a slightly lower average test MAE than the target-regime gate. It shares the same NWP-driven base predictions and changes only the gate context. It therefore does not measure the removal of all future weather information. Nor is it promoted as a new winning configuration after observing these test results. Together with the confirmation failures and the bootstrap interval, this ablation indicates that the additional weather-conditioning effect is weak in the present public setting.

\begin{figure}[t]
\centering
\includegraphics[width=\columnwidth]{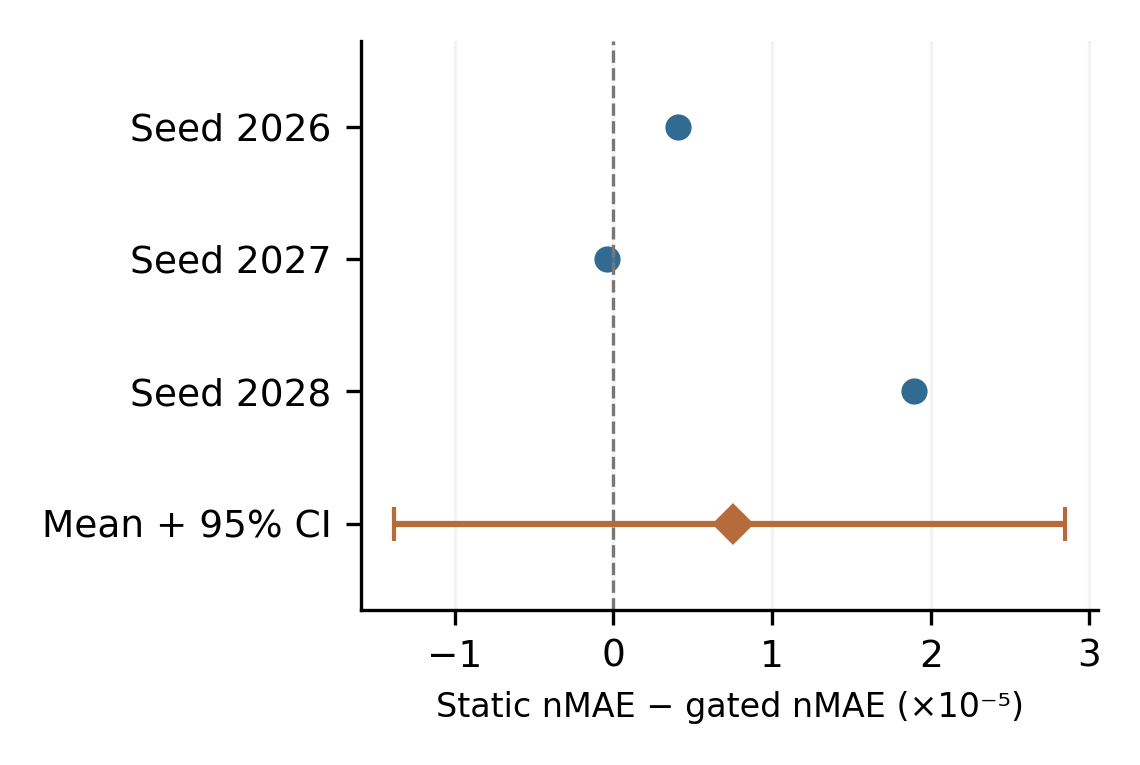}
\caption{Incremental gate gain relative to static ensembling. Positive values indicate lower gate nMAE. Individual seed gains and the paired calendar-day bootstrap interval show that the small aggregate improvement does not establish a stable benefit.}
\label{fig:gate}
\end{figure}

\subsection{Additional datasets and retrospective weather condition}
Three further public sources extend the evaluation while retaining the V9 combination mechanism. The Ausgrid release supplies residential generation in Australia \citep{ausgridSolarData,ratnam2017ausgrid}; the OPSD household release supplies original German electricity-meter feeds \citep{opsd2020household,wiese2019opsd}; and PVDAQ supplies measured AC power from system 34 in Las Vegas \citep{deline2021pvdaq}. These are distinct data sources, rather than partitions of the GEFCom mirror. Site selection uses source identity and data quality before model scoring: three eligible Ausgrid households, three OPSD households with adequate common coverage, and the specified PVDAQ installation. This is temporal evaluation within the selected sites, not a held-out-site transfer experiment.

\begin{table*}[t]
\centering
\small
\caption{Dataset coverage and disjoint chronological evaluation stages.}
\label{tab:additional_datasets}
\begin{tabular*}{\textwidth}{@{\extracolsep{\fill}}lccc@{}}
\toprule
Method / quantity & PVDAQ & OPSD & Ausgrid \\
\midrule
Stations & 1 & 3 & 3 \\
Hourly end-label range (UTC) & \shortstack{2011-01-01 09:00\\to 2014-01-01 08:00} & \shortstack{2016-03-01 01:00\\to 2018-01-01 00:00} & \shortstack{2011-06-30 15:00\\to 2013-06-30 14:00} \\
Valid / total station-hours & 25,405/26,304 (96.58\%) & 46,073/48,312 (95.37\%) & 52,608/52,632 (99.95\%) \\
Train windows & 4,343 & 9,217 & 11,955 \\
Checkpoint windows & 1,040 & 1,596 & 2,406 \\
Weight Fit windows & 1,031 & 1,712 & 2,376 \\
Selection windows & 605 & 976 & 1,188 \\
Confirmation windows & 526 & 979 & 1,218 \\
Test windows & 2,329 & 3,516 & 4,785 \\
Test target points & 9,316 & 14,064 & 19,140 \\
Test target-label range (UTC) & \shortstack{2013-05-27 14:00\\to 2014-01-01 04:00} & \shortstack{2017-08-20 05:00\\to 2017-12-31 19:00} & \shortstack{2013-02-04 20:00\\to 2013-06-30 10:00} \\
\bottomrule
\end{tabular*}
\par\smallskip
\begin{minipage}{\textwidth}\footnotesize All tasks use 24 hourly historical values and four subsequent hourly targets; complete target interval midpoints must be in local [06:00,20:00). The grid retains missing hours; windows containing missing input or target values are excluded. Raw valid-hour counts include signed observations where retained by the source audit. Target stages are disjoint, whereas rolling windows within a stage overlap. The hourly end-label range uses inclusive first and last labels. Normalization uses each station's training maximum, not installed capacity; weather inputs are retrospective ERA5.\end{minipage}
\end{table*}


Ausgrid half-hourly generation energy is converted to hourly mean power only when both measurements are present and the source marks the day as measured. Ambiguous or nonexistent daylight-saving timestamps are excluded. OPSD labels are reconstructed from original cumulative kWh readings, rather than its interpolated hourly release. The energy difference is divided by the actual endpoint duration; endpoints must precede their nominal hour boundaries by no more than 180 seconds, and internal observation gaps above 300 seconds or counter anomalies invalidate the label. Consequently, these are approximately hourly averages whose actual endpoints remain available for audit. PVDAQ uses the official power-unit conversion and requires four unique, finite observations at exact quarter-hour clock times. Signed AC readings, including small negative values, are retained. Its instrument averaging alignment and historical daylight-saving behavior are not independently certified.

All three sources are paired with ERA5 reanalysis \citep{hersbach2020era5,c3s2018era5data} retrieved through Open-Meteo with the ERA5 model explicitly requested \citep{zippenfenig2024openmeteo}. Eight covariates comprise shortwave radiation, cloud cover, precipitation, temperature, relative humidity, surface pressure, wind speed, and wind direction. Radiation is the preceding-hour mean and precipitation the preceding-hour total; the other fields are instantaneous values. UTC weather labels are aligned to nominal power-interval ends. Ausgrid uses postcode reference coordinates and OPSD a common Konstanz regional point; neither represents a verified rooftop sensor location. PVDAQ uses its published site coordinates. Requests, returned grids, elevation settings, units, responses, and checksums are archived.

\emph{These additional experiments are retrospective ERA5-assisted evaluations.} Target-period reanalysis is supplied equally to the neural models and complete-feature regression controls, but it is unavailable as such at a real forecast origin. The results therefore test the retained architecture under a disclosed oracle-weather condition; they do not validate an operational NWP forecast pipeline. The observed PVDAQ weather is retained in the source data but is not silently substituted for ERA5 or presented as a forecast.

\subsection{Additional protocol and matched controls}
Each dataset uses the same 24-hour history and four-hour target duration. All 28 power observations must be present on the hourly UTC grid; no missing history or target is interpolated. Only the four target interval midpoints must lie in the local clock range $[06{:}00,20{:}00)$. A first target ending at $T$ corresponds to an issue time of $T-1$ hour. The common time range is divided chronologically into 50\% base training, 10\% checkpoint selection, 10\% weight fitting, 5\% parameter selection, 5\% confirmation, and 20\% test. Complete target intervals must stay within their partition. Earlier observations can enter later rolling histories.

For these sources, $c_s$ is the maximum valid power within the first training partition. Errors normalized by this training scale are denoted sMAE and sRMSE, distinguishing them from the capacity-normalized GEFCom metrics. Neither the denominator nor weather standardization uses later partitions. Labels remain unmodified, while predictions from every constituent and combination are clipped to $[0,1.25]$. Physical-unit MAE and RMSE, training scales, and per-site and per-horizon errors are also archived.

The original public adaptation's nine neural families, five classical constituents, hyperparameters, three seeds, and 12-epoch budget are retained. Weather dimensionality changes from 12 to eight, and site embeddings are fitted anew. The bank's ridge and gradient-boosting constituents are trained only on the base-training partition, ensuring that their weight-fitting predictions are out of sample. Both now receive the complete flattened historical power/weather, future weather/calendar, and site indicators. Separate matched controls select ridge penalty from $\{0.1,1,10,100\}$ or boosting iterations from $\{75,150,300\}$ using the checkpoint-selection partition, then refit on the union of base training and the three ensemble fitting/selection/confirmation partitions. They thereby receive the labels accessible to the final refitted ensemble. These controls are not inserted into the ensemble bank. Official-author DLinear code \citep{zeng2023}, with the same neural epoch budget, provides an additional historical-power-only reference; it does not share the meteorological information set.

The original simplex solvers, shrinkage candidates, bounded gate, regularization candidates, and confirmation rule are unchanged. The empirical radiation envelope uses only training-period monthly/hourly values. Cloud cover is already in percent and precipitation in millimetres, requiring no GEFCom-specific unit multiplier. Confirmation is divided into three chronological folds; the candidate gate must improve in every fold. After the original prescribed refitting, all candidates and the confirmation-selected fallback are scored on the common test set.

Two primary contrasts are fixed for each dataset: static fusion against matched boosting, and gated against static fusion. Losses are averaged across seeds before 10,000 circular seven-calendar-day bootstrap replicates, keeping all sites and overlapping horizons together. We report descriptive 95\% intervals and Bonferroni intervals for these six primary contrasts. Additional ablation contrasts are exploratory. All intervals are conditional on these sites, temporal splits, and fitted runs.

\subsection{Component ablations}
The ablations share the same frozen bank and final ensemble-fitting labels. Removing the gate yields the static ensemble. Removing shrinkage retains the fitted local station--horizon weights; removing site dependence uses only global horizon weights; equal weighting removes weight fitting altogether. A final variant supplies the last historical regime to the gate over every horizon, retaining the full gate's selected regularization and the same ERA5-informed base predictions. It isolates gate context, not removal of all future weather information. Individual bank members provide complementary component comparisons.

PVDAQ contains one site. Its local and global objectives coincide, so the shrinkage and site-dependence ablations are degenerate there and cannot establish a cross-site benefit. Those mechanisms must be assessed on the two multi-household datasets.


\subsection{Additional comparison and ablation results}
\begin{table*}[t]
\centering
\small
\caption{Additional-data comparison of the V9-based public-data adaptation.}
\label{tab:additional_main}
\begin{tabular*}{\textwidth}{@{\extracolsep{\fill}}lccc@{}}
\toprule
Method / quantity & PVDAQ & OPSD & Ausgrid \\
\midrule
Gradient boosting (matched) & $0.040704 \pm 0.000000$ & $0.059155 \pm 0.000000$ & $0.067875 \pm 0.000000$ \\
Ridge (matched) & $0.047269 \pm 0.000000$ & $0.068594 \pm 0.000000$ & $0.080067 \pm 0.000000$ \\
Static ensemble & $0.040252 \pm 0.000178$ & $0.056548 \pm 0.000377$ & $0.066767 \pm 0.000285$ \\
V9 regime gate & $0.040137 \pm 0.000269$ & $0.056617 \pm 0.000442$ & $0.066724 \pm 0.000304$ \\
Confirmation-selected rule & $0.040252 \pm 0.000178$ & $0.056547 \pm 0.000377$ & $0.066767 \pm 0.000285$ \\
\bottomrule
\end{tabular*}
\par\smallskip
\begin{minipage}{\textwidth}\footnotesize Entries are test sMAE mean $\pm$ sample standard deviation across three training seeds. Each station is normalized by its training-period maximum; sMAE is not capacity-normalized nMAE. Lower is better. Future ERA5 is a retrospective oracle input, not an operational weather forecast. Matched Ridge/GB use the same complete feature information and final fitting-label access as the ensemble.\end{minipage}
\end{table*}



The three ERA5-assisted benchmarks contain 10,630 test windows and 42,520 window--horizon targets. These counts refer to the shared test population, not the number of training seeds. Static fusion reduces sMAE relative to matched gradient boosting by 1.11\% on PVDAQ, 4.41\% on OPSD, and 1.63\% on Ausgrid (Table~\ref{tab:additional_main}).

The strength of this evidence differs across datasets. After correction for the six primary comparisons, only the OPSD static-fusion contrast excludes zero. The lower mean errors on PVDAQ and Ausgrid therefore do not establish the same level of support. Gating provides no supported incremental improvement on any of the three datasets. It passes the three-fold confirmation rule in none of the PVDAQ or Ausgrid runs and in one of the three OPSD runs. Table~\ref{tab:additional_main} reports both the candidate gate and the rule that falls back to static fusion when confirmation fails.

\begin{figure*}[t]
\centering
\includegraphics[width=\textwidth]{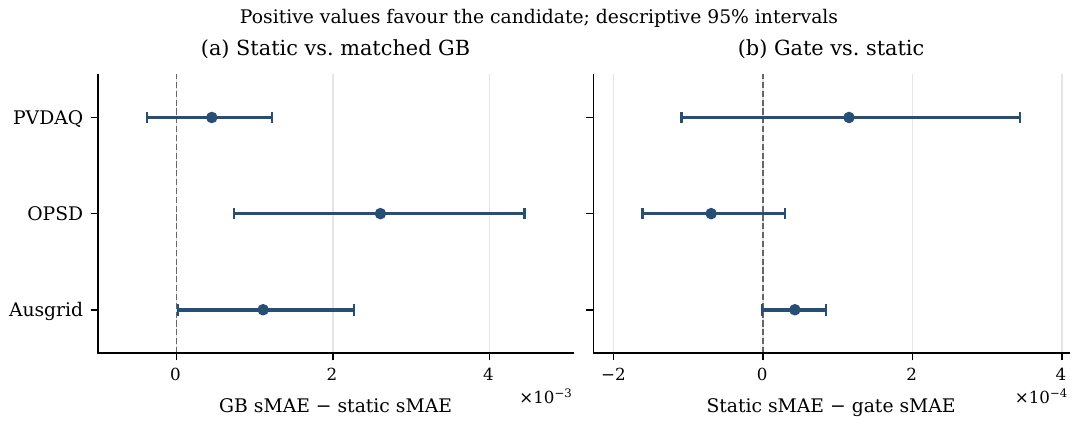}
\caption{Paired effects on the additional datasets. Positive values indicate reduced sMAE for static fusion against matched boosting (left), or for gating against static fusion (right). Bars are descriptive 95\% circular seven-day bootstrap intervals, conditional on the three fitted seeds. The separate six-primary-contrast correction is used for the inferential statements in the text. All panels use retrospective ERA5 covariates and training-maximum normalization.}
\label{fig:additional_effects}
\end{figure*}

\begin{table*}[t]
\centering
\small
\caption{Component ablations for the V9-based public-data adaptation.}
\label{tab:additional_ablations}
\begin{tabular*}{\textwidth}{@{\extracolsep{\fill}}lccc@{}}
\toprule
Method / quantity & PVDAQ & OPSD & Ausgrid \\
\midrule
V9 regime gate & $0.040137 \pm 0.000269$ & $0.056617 \pm 0.000442$ & $0.066724 \pm 0.000304$ \\
Static ensemble & $0.040252 \pm 0.000178$ & $0.056548 \pm 0.000377$ & $0.066767 \pm 0.000285$ \\
Station weights, no shrinkage & $0.040252 \pm 0.000178$ & $0.056964 \pm 0.000425$ & $0.066917 \pm 0.000252$ \\
Global horizon weights & $0.040252 \pm 0.000178$ & $0.056536 \pm 0.000343$ & $0.066717 \pm 0.000237$ \\
Uniform ensemble & $0.050021 \pm 0.000036$ & $0.066052 \pm 0.000311$ & $0.076001 \pm 0.000301$ \\
Origin-context gate & $0.040324 \pm 0.000199$ & $0.056607 \pm 0.000443$ & $0.066780 \pm 0.000268$ \\
\bottomrule
\end{tabular*}
\par\smallskip
\begin{minipage}{\textwidth}\footnotesize Test sMAE mean $\pm$ three-seed sample standard deviation, using training-maximum normalization. All variants share the frozen predictor bank. Static removes the regime gate; no shrinkage removes the global-weight shrinkage; global horizon weights remove station-specific weighting; uniform uses equal weights. Origin-context gating changes only the gate context and retains oracle-informed base predictions and the main gate's selected regularization. It is not a full causal-input ablation. These rows are secondary comparisons; the six-primary-contrast family-wise correction must not be generalized to them.\end{minipage}
\end{table*}


Weight learning accounts for a clearer change than weather gating (Table~\ref{tab:additional_ablations}). Learned static weights lower mean error relative to uniform averaging on all three datasets. On OPSD, removing shrinkage increases sMAE from 0.056548 to 0.056964, whereas shared global horizon weights give 0.056536. The corresponding Ausgrid errors are 0.066767, 0.066917, and 0.066717. Local weights alone thus perform worse than either shrinkage or global sharing in these two samples. The small differences between shrinkage and global sharing provide little reason to prefer the more site-specific estimate. These comparisons are exploratory; the six-test correction applies only to the primary contrasts. PVDAQ's identical local and global rows follow from its single-site design.

Changing the gate context produces smaller and less consistent effects. The target-period context uses an oracle, while the origin-context variant changes only the gate and retains the same oracle-informed base forecasts. This comparison cannot establish the performance of a system restricted to information available in real time. Table~\ref{tab:additional_full} reports every model's sMAE; physical-unit, per-site, and per-horizon results are provided with the experiment records.


\subsection{Dependence on individual members and groups}
\begin{table}[t]
\centering
\caption{Fixed-weight group removal: mean sMAE over three seeds. Removing either group worsens every seed on each dataset. These are exploratory test diagnostics without refitting.}
\label{tab:member_groups}
\small
\begin{tabular}{lrrr}
\toprule
Retained forecasts & PVDAQ & OPSD & Ausgrid \\
\midrule
Full static (14) & 0.040252 & 0.056548 & 0.066767 \\
Classical only (5) & 0.041669 & 0.059649 & 0.070592 \\
Neural only (9) & 0.045201 & 0.059391 & 0.068401 \\
\bottomrule
\end{tabular}
\end{table}


Removing all neural members or all classical members increases error on each added dataset (Table~\ref{tab:member_groups}). Every group-removal contrast worsens in all three seeds. The mean sMAE increases range from 0.001417 to 0.004949. Thus, both groups contribute to the particular frozen mixture, even though no individual neural model consistently outperforms the matched boosting control.

This group-level result does not imply that all fourteen constituents are indispensable. The bank's gradient-boosting member has the largest mean single-removal deterioration on each dataset: 0.003721, 0.001827, and 0.001787 on PVDAQ, OPSD, and Ausgrid. In contrast, removing TiDE reduces OPSD sMAE by 0.000177, and removing persistence reduces Ausgrid sMAE by 0.000171. These are small, descriptive changes, not separately tested discoveries. Complete single-member results appear in Table~\ref{tab:member_removal}.

Across three datasets, three seeds, and seventeen scenarios, 153 diagnostic evaluations were reconstructed from frozen predictions. An independent complementary-sum implementation agrees in sMAE to within $2.8\times10^{-17}$, and all 27 source-file hashes remain unchanged. This verifies the calculation, not the independence of the already inspected test sets. No new member selection or training follows from these results.


\subsection{Secondary case: a native member replacement}
\label{sec:native}
The native case examines a separate engineering implementation denoted V28. It combines TiDE, PatchTST, LSTM, and HGB16, a histogram-gradient-boosting predictor with sixteen separately fitted horizon heads. HGB16 occupies one fusion slot, so four slots do not mean four equally sized models. The matched control retains the same three neural checkpoints and substitutes the original Transformer for HGB16. Both pipelines use the same development procedure: static convex fusion with regularization 0.1, followed by additive global-horizon and station-intercept median calibration. Horizon residual medians use shrinkage prior 256; station medians, pooled over horizons after the horizon correction, use prior 128. Each term is clipped at $\pm0.03$ normalized units before shrinkage; their sum is clipped at the same bound, and final forecasts at $[0,1.25]$. This is not the V9 weather gate.

Each window contains 96 historical fifteen-minute intervals and predicts sixteen subsequent interval-average powers. The historical and future feature arrays have 33 and 45 channels, respectively, including fifteen weather channels and time and availability features. The feed uses an assumed NWP availability delay of 370 minutes after its run time; independently verified operational arrival records are unavailable. The task-specific compact neural backbones use one training run under protocol seed 20260913, fixed model-specific initialization seeds, twenty epochs, and 125,776 training windows. Each HGB head uses 187 summary features, absolute-error loss, 400 boosting iterations, learning rate 0.05, at most 31 leaves, minimum leaf size 64, and $L_2$ regularization 1. These are matched-input pipeline comparisons, not matched total training budgets.

Frozen predictions are evaluated on 24 sites, 16,114 common windows, and 257,824 window--horizon targets from 1--18 July 2026. Both combinations use an identical valid target mask and the same engineering capacity-reference snapshot for nMAE; the snapshot's nameplate provenance is not independently certified. This period was inspected during prior development; the analysis adds no new training or untouched test data. Paired bootstrap intervals use 20,000 circular three-day resamples, retaining sites and horizons together, with seven-day sensitivity analysis. These intervals describe the fixed historical cohort and cannot remove selection bias.

\begin{table}[t]
\centering
\caption{Native fifteen-minute case on previously inspected private targets. The first pair shares the complete fitting and calibration procedure. Remaining rows omit calibration and are diagnostic comparisons.}
\label{tab:native_accuracy}
\small
\begin{tabular}{lr}
\toprule
Configuration & nMAE \\
\midrule
Four neural, calibrated & 0.076767 \\
V28 replacement, calibrated & 0.075446 \\
\midrule
Four neural, uncalibrated & 0.077431 \\
V28 replacement, uncalibrated & 0.076057 \\
V28 uniform, uncalibrated & 0.076254 \\
HGB16 alone & 0.078624 \\
\bottomrule
\end{tabular}
\end{table}


The calibrated replacement reduces nMAE by 1.72\% relative to the matched four-neural pipeline. The transformed score $100(1-\mathrm{nMAE})$ rises from 92.323336\% to 92.455413\%, a gain of 0.132077 percentage points. Error falls at 22 sites and increases at two; every horizon aggregate improves. The descriptive 95\% interval for the score gain is $[0.05868,0.20675]$ percentage points; the three-contrast adjusted interval is $[0.04289,0.22231]$. Seven-day resampling also retains a positive interval. Eighteen days and one seed do not establish long-term stability. The improvement is also present without calibration, while uniform fusion and HGB16 alone are exploratory, uncalibrated references. An older system archive, V18, has lower same-target nMAE (0.072125), but different training and calibration prevent using it as a controlled component contrast. Thus, the replacement has the better score in the matched pair, not the highest score across all archived systems.

\subsection{Measured inference cost of the replacement}
\begin{table}[t]
\centering
\caption{Measured inference and saved checkpoints. Latencies are medians in ms per window; batch-256 times are amortized. The two pipelines share hardware and timed inputs.}
\label{tab:native_cost}
\small
\begin{tabular}{lrr}
\toprule
Quantity & Four neural & V28 \\
\midrule
Batch 1 (ms/window) & 14.421 & 69.597 \\
Batch 256 (ms/window) & 0.0922 & 0.6533 \\
Checkpoint files (bytes) & 5,880,746 & 15,939,797 \\
\bottomrule
\end{tabular}
\end{table}


Both complete combinations are timed on the same RTX 3060 Laptop GPU and host CPU, with two CPU threads, Python 3.12.13, PyTorch 2.11.0+cu128, and scikit-learn 1.9.0. The fixed sample comprises the first 256 checkpoint-selection windows, all from one site. Two warm-ups precede seven alternating repetitions. Timings include device copies, neural forwards, HGB feature construction and all sixteen heads, fusion, and calibration; they exclude loading, data retrieval, network service, and training. Batch-one and batch-256 outputs agree with frozen predictions within $2\times10^{-6}$.

Table~\ref{tab:native_cost} shows that the replacement is slower in this implementation. The median paired latency ratios are 4.71 for individual windows and 7.96 for batches of 256; these are medians of paired ratios, not ratios of the table's median times. Saved checkpoint bytes also increase. File size includes serialization overhead and is not a measurement of RAM or GPU memory. The timing sample and hardware limit extrapolation to a multi-site server, and no energy or end-to-end service measurement is claimed.


\subsection{Discussion and limits of the evidence}
The first question has a qualified positive answer. Learned static fusion lowers mean error relative to uniform averaging on the added datasets, but its advantage over a complete-feature, matched-label boosting control survives the six-comparison correction only on OPSD. The original GEFCom difference is less controlled because its regression baseline has fewer inputs and fitting labels. Neither comparison establishes that a large bank is always preferable to a well-tuned single predictor.

For the second question, the added weather gate has no consistent incremental support. Global sharing performs close to or slightly better than local shrinkage in the two multi-site additions, while group removals expose dependence on both neural and classical predictions in the current frozen combination. Some individual removals nevertheless reduce error. These observations distinguish the contribution of a group from the necessity of every member. A smaller retrained bank might recover the same information with different weights; the present removal analysis cannot answer that question.

The native replacement case answers the third question for one implementation and one previously inspected period. A tree-based member reduces error relative to the matched four-neural pipeline, including before calibration, while increasing measured inference latency and saved checkpoint size. A 69.6~ms single-window latency may still be compatible with a fifteen-minute issue interval. No service deadline, request volume, or monetary value of forecast error is specified, so the measured tradeoff does not by itself establish deployment value. Nor were energy use, memory consumption, or total training and search costs measured.

Several limits apply across the study. The public tests use one temporal holdout per dataset and three seeds, with fixed-budget, task-adapted architectures. Matching input and fitting-label access does not match all optimization opportunities or total compute. The added sources use training-maximum scaling, regional ERA5, and source-specific time and measurement assumptions. Their target-period reanalysis is a retrospective oracle; the GEFCom mirror also lacks forecast-issuance metadata. The private archives have further normalization and development-history limits. The 82-member archive, fourteen-member hourly study, and native four-slot case therefore do not form a common accuracy--complexity curve.

Finally, numerical reproducibility is distinct from independence of evaluation. Hashes and independent recomputation support the reported numbers, but do not undo repeated inspection of a historical test set. The new removals are exploratory and the native bootstrap intervals are conditional on a reused period. Establishing a deployable reduced ensemble requires freezing member selection, calibration, and compute budgets using development data, then evaluating genuinely unexposed dates with verifiable forecast vintages. No such prospective validation is claimed here.

\section{Conclusion}
This study assessed the contribution of ensemble components to four-hour PV forecasting. Its primary public implementation retains the V9 weighting mechanism in a fourteen-member hourly bank, while treating the original 82-member result as an archive audit. Static fusion has corrected positive evidence against matched boosting on OPSD; weather gating does not show a consistent additional benefit. Exploratory removals reveal dependence on both neural and classical groups, without establishing that all fourteen members are necessary.

A separate native fifteen-minute comparison finds a 1.72\% relative nMAE reduction after one member replacement, accompanied by slower inference and larger saved checkpoints. Together, these results support reporting incremental predictive effects and measured costs, rather than using ensemble size or a transformed accuracy score as evidence of value. The weather-oracle condition, limited temporal holdouts, and historical test reuse require further validation before claims of operational generalization or optimal ensemble size can be made.

\section*{Data and code availability}
The public sources comprise the GEFCom2014 researcher mirror, Ausgrid generation data, original OPSD household feeds, and NREL PVDAQ system 34. Additional weather is ERA5 served through Open-Meteo. Download scripts preserve attribution, request parameters, response hashes, and preprocessing rules. The existing public reproducibility packages retain model snapshots, trained checkpoints, weights, predictions, ablations, and numerical audits. A separate revision supplement adds the fixed-weight removal analysis, its source hashes, and summarized native accuracy and timing evidence. It depends on the existing public prediction caches and is not a replacement for those packages. Private raw data, predictions, and checkpoints are not redistributed. The native case consequently has auditable local evidence but is not fully reproducible from the public supplement alone.

\appendix
\section{Reproduction and evidence boundaries}
The public runner is \texttt{run\_public\_benchmark.py}; its default settings define the reported three-seed protocol. Result tables are generated from the saved JSON and NPZ outputs. Validation choices are saved before test-loss evaluation. Restarting a completed run loads its stored results instead of searching for a new configuration. A changed protocol should use a new experiment directory and a new evaluation plan.

The retained private gate uses ordinary row-disjoint chronological subdivisions of its validation windows. The source split assertion checks window-row membership, not the independence of all target timestamps across these internal subdivisions. Public subdivisions additionally require complete target intervals to remain within their own boundaries. The private weights' simplex constraints and zero-gate identity are verified, but these numerical checks do not certify every historical training decision, label-cleaning rule, or meteorological input vintage.

\begin{table*}[p]
\centering
\small
\caption{Complete model comparison on the three additional datasets.}
\label{tab:additional_full}
\begin{tabular*}{\textwidth}{@{\extracolsep{\fill}}lccc@{}}
\toprule
Method / quantity & PVDAQ & OPSD & Ausgrid \\
\midrule
Persistence & $0.256179 \pm 0.000000$ & $0.165981 \pm 0.000000$ & $0.252482 \pm 0.000000$ \\
Daily persistence & $0.065706 \pm 0.000000$ & $0.112617 \pm 0.000000$ & $0.126343 \pm 0.000000$ \\
Training climatology & $0.116098 \pm 0.000000$ & $0.169743 \pm 0.000000$ & $0.154308 \pm 0.000000$ \\
Ridge (bank) & $0.048918 \pm 0.000000$ & $0.068763 \pm 0.000000$ & $0.079835 \pm 0.000000$ \\
Gradient boosting (bank) & $0.041994 \pm 0.000000$ & $0.058105 \pm 0.000000$ & $0.068300 \pm 0.000000$ \\
Ridge (matched) & $0.047269 \pm 0.000000$ & $0.068594 \pm 0.000000$ & $0.080067 \pm 0.000000$ \\
Gradient boosting (matched) & $0.040704 \pm 0.000000$ & $0.059155 \pm 0.000000$ & $0.067875 \pm 0.000000$ \\
LSTM & $0.051412 \pm 0.001468$ & $0.067351 \pm 0.001325$ & $0.075920 \pm 0.000779$ \\
TimeMixer & $0.048119 \pm 0.001806$ & $0.065182 \pm 0.001800$ & $0.071528 \pm 0.000447$ \\
Transformer & $0.047463 \pm 0.001601$ & $0.064353 \pm 0.001389$ & $0.073492 \pm 0.000941$ \\
Mamba-style & $0.048958 \pm 0.002844$ & $0.062884 \pm 0.000109$ & $0.070929 \pm 0.000800$ \\
PatchTST & $0.050813 \pm 0.001139$ & $0.070003 \pm 0.002846$ & $0.078176 \pm 0.001183$ \\
TiDE & $0.050237 \pm 0.001101$ & $0.065087 \pm 0.000910$ & $0.071214 \pm 0.000169$ \\
TSMixer & $0.049436 \pm 0.000661$ & $0.062828 \pm 0.002023$ & $0.071849 \pm 0.002287$ \\
ModernTCN & $0.047129 \pm 0.000710$ & $0.063006 \pm 0.002109$ & $0.070791 \pm 0.001284$ \\
TFT & $0.049464 \pm 0.002027$ & $0.063557 \pm 0.002042$ & $0.072522 \pm 0.000544$ \\
DLinear (power only) & $0.066435 \pm 0.000778$ & $0.092685 \pm 0.000362$ & $0.104919 \pm 0.000363$ \\
Uniform ensemble & $0.050021 \pm 0.000036$ & $0.066052 \pm 0.000311$ & $0.076001 \pm 0.000301$ \\
Global horizon weights & $0.040252 \pm 0.000178$ & $0.056536 \pm 0.000343$ & $0.066717 \pm 0.000237$ \\
Station weights, no shrinkage & $0.040252 \pm 0.000178$ & $0.056964 \pm 0.000425$ & $0.066917 \pm 0.000252$ \\
Static ensemble & $0.040252 \pm 0.000178$ & $0.056548 \pm 0.000377$ & $0.066767 \pm 0.000285$ \\
Origin-context gate & $0.040324 \pm 0.000199$ & $0.056607 \pm 0.000443$ & $0.066780 \pm 0.000268$ \\
V9 regime gate & $0.040137 \pm 0.000269$ & $0.056617 \pm 0.000442$ & $0.066724 \pm 0.000304$ \\
Confirmation-selected rule & $0.040252 \pm 0.000178$ & $0.056547 \pm 0.000377$ & $0.066767 \pm 0.000285$ \\
\bottomrule
\end{tabular*}
\par\smallskip
\begin{minipage}{\textwidth}\footnotesize Entries are test sMAE mean $\pm$ sample standard deviation across three training seeds. Each station is normalized by its training-period maximum; sMAE is not capacity-normalized nMAE. Lower is better. Future ERA5 is a retrospective oracle input, not an operational weather forecast. DLinear uses historical power only; its information set differs from the weather-assisted models. The nine neural rows are original-repository model families retrained for the hourly task. Bank Ridge/GB use the initial training stage; matched controls additionally fit the ensemble's meta-stage labels. A zero standard deviation for deterministic controls does not mean zero statistical uncertainty.\end{minipage}
\end{table*}


\section{Complete member-removal diagnostics}
Table~\ref{tab:member_removal} reports every fixed-weight removal. Differences are descriptive three-seed means on already inspected test periods. Entries displayed as approximately zero have absolute magnitude below $0.0005$ in the table's scaled units. No member is removed from the retained system on this basis.
\begin{table}[pos=!htbp]
\centering
\caption{Exploratory fixed-weight removal sensitivity on previously inspected test sets. Values are mean changes in sMAE multiplied by $10^{3}$; positive values indicate deterioration after removal. Remaining station--horizon weights are renormalized without refitting.}
\label{tab:member_removal}
\small
\begin{tabular}{lrrr}
\toprule
Removed member(s) & PVDAQ & OPSD & Ausgrid \\
\midrule
LSTM & +0.006 & +0.001 & $\approx0$ \\
TimeMixer & +0.026 & -0.003 & +0.189 \\
Transformer & +0.116 & +0.132 & -0.004 \\
Mamba & $\approx0$ & -0.095 & +0.015 \\
PatchTST & +0.073 & +0.135 & +0.015 \\
TiDE & $\approx0$ & -0.177 & +0.119 \\
TSMixer & +0.078 & +0.179 & +0.089 \\
ModernTCN & -0.011 & -0.007 & +0.029 \\
TFT & +0.197 & +0.366 & -0.033 \\
Persistence & +0.002 & -0.021 & -0.171 \\
Daily persistence & $\approx0$ & -0.003 & $\approx0$ \\
Climatology & $\approx0$ & -0.004 & -0.001 \\
Ridge & +0.135 & -0.022 & -0.005 \\
GB (bank) & +3.721 & +1.827 & +1.787 \\
All neural (9) & +1.417 & +3.101 & +3.825 \\
All classical (5) & +4.949 & +2.843 & +1.634 \\
\bottomrule
\end{tabular}
\end{table}


\FloatBarrier


\bibliographystyle{cas-model2-names}
\renewcommand{\bibfont}{\fontsize{8pt}{9.5pt}\selectfont}
\bibliography{references}

@article{antonanzas2016,
  title = {Review of photovoltaic power forecasting},
  author = {Antonanzas, J. and Osorio, N. and Escobar, R. and Urraca, R. and Martinez-de-Pison, F.J. and Antonanzas-Torres, F.},
  year = {2016},
  journal = {Solar Energy},
  volume = {136},
  pages = {78-111},
  doi = {10.1016/j.solener.2016.06.069},
}

@article{hong2016,
  title = {Probabilistic energy forecasting: Global Energy Forecasting Competition 2014 and beyond},
  author = {Hong, Tao and Pinson, Pierre and Fan, Shu and Zareipour, Hamidreza and Troccoli, Alberto and Hyndman, Rob J.},
  year = {2016},
  journal = {International Journal of Forecasting},
  volume = {32},
  number = {3},
  pages = {896-913},
  doi = {10.1016/j.ijforecast.2016.02.001},
}

@article{dumas2022,
  title = {A deep generative model for probabilistic energy forecasting in power systems: normalizing flows},
  author = {Dumas, Jonathan and Wehenkel, Antoine and Lanaspeze, Damien and Corn{\'e}lusse, Bertrand and Sutera, Antonio},
  year = {2022},
  journal = {Applied Energy},
  volume = {305},
  pages = {117871},
  doi = {10.1016/j.apenergy.2021.117871},
}

@article{tft2021,
  title = {Temporal Fusion Transformers for interpretable multi-horizon time series forecasting},
  author = {Lim, Bryan and Ar{\i}k, Sercan {\"O}. and Loeff, Nicolas and Pfister, Tomas},
  year = {2021},
  journal = {International Journal of Forecasting},
  volume = {37},
  number = {4},
  pages = {1748-1764},
  doi = {10.1016/j.ijforecast.2021.03.012},
}

@article{hewamalage2023,
  title = {Forecast evaluation for data scientists: common pitfalls and best practices},
  author = {Hewamalage, Hansika and Ackermann, Klaus and Bergmeir, Christoph},
  year = {2023},
  journal = {Data Mining and Knowledge Discovery},
  volume = {37},
  number = {2},
  pages = {788-832},
  doi = {10.1007/s10618-022-00894-5},
}

@article{wang2023,
  title = {Forecast combinations: An over 50-year review},
  author = {Wang, Xiaoqian and Hyndman, Rob J. and Li, Feng and Kang, Yanfei},
  year = {2023},
  journal = {International Journal of Forecasting},
  volume = {39},
  number = {4},
  pages = {1518-1547},
  doi = {10.1016/j.ijforecast.2022.11.005},
}

@article{hochreiter1997,
  title = {Long Short-Term Memory},
  author = {Hochreiter, Sepp and Schmidhuber, J{\"u}rgen},
  year = {1997},
  journal = {Neural Computation},
  volume = {9},
  number = {8},
  pages = {1735-1780},
  doi = {10.1162/neco.1997.9.8.1735},
}

@article{friedman2001,
  title = {Greedy function approximation: A gradient boosting machine.},
  author = {Friedman, Jerome H.},
  year = {2001},
  journal = {The Annals of Statistics},
  volume = {29},
  number = {5},
  doi = {10.1214/aos/1013203451},
}

@article{zeng2023,
  title = {Are Transformers Effective for Time Series Forecasting?},
  author = {Zeng, Ailing and Chen, Muxi and Zhang, Lei and Xu, Qiang},
  year = {2023},
  journal = {Proceedings of the AAAI Conference on Artificial Intelligence},
  volume = {37},
  number = {9},
  pages = {11121-11128},
  doi = {10.1609/aaai.v37i9.26317},
}

@article{weatherensemble2020,
  title = {Ensemble models for solar power forecasting---a weather classification approach},
  author = {Amarasinghe, P. A. G. M. and Abeygunawardana, N. S. and Jayasekara, T. N. and Edirisinghe, E. A. J. P. and Abeygunawardane, S. K.},
  year = {2020},
  journal = {AIMS Energy},
  volume = {8},
  number = {2},
  pages = {252-271},
  doi = {10.3934/energy.2020.2.252},
}

@article{weathermars2019,
  title = {Weather-Classification-MARS-Based Photovoltaic Power Forecasting for Energy Imbalance Market},
  author = {Zhang, Xiaoning and Fang, Fang and Liu, Jizhen},
  year = {2019},
  journal = {IEEE Transactions on Industrial Electronics},
  volume = {66},
  number = {11},
  pages = {8692-8702},
  doi = {10.1109/TIE.2018.2889611},
}

@article{weatherchain2023,
  title = {Pairing ensemble numerical weather prediction with ensemble physical model chain for probabilistic photovoltaic power forecasting},
  author = {Mayer, Martin J{\'a}nos and Yang, Dazhi},
  year = {2023},
  journal = {Renewable and Sustainable Energy Reviews},
  volume = {175},
  pages = {113171},
  doi = {10.1016/j.rser.2023.113171},
}

@article{pvreview2022,
  title = {Solar Photovoltaic Power Forecasting: A Review},
  author = {Iheanetu, Kelachukwu J.},
  year = {2022},
  journal = {Sustainability},
  volume = {14},
  number = {24},
  pages = {17005},
  doi = {10.3390/su142417005},
}

@article{deepreview2024,
  title = {Towards energy efficiency: A comprehensive review of deep learning-based photovoltaic power forecasting strategies},
  author = {Husein, Mauladdawilah and Gago, E.J. and Hasan, Balfaqih and Pegalajar, M.C.},
  year = {2024},
  journal = {Heliyon},
  volume = {10},
  number = {13},
  pages = {e33419},
  doi = {10.1016/j.heliyon.2024.e33419},
}

@article{nwpdataset2022,
  title = {An archived dataset from the ECMWF Ensemble Prediction System for probabilistic solar power forecasting},
  author = {Wang, Wenting and Yang, Dazhi and Hong, Tao and Kleissl, Jan},
  year = {2022},
  journal = {Solar Energy},
  volume = {248},
  pages = {64-75},
  doi = {10.1016/j.solener.2022.10.062},
}

@article{jacobs1991,
  title = {Adaptive Mixtures of Local Experts},
  author = {Jacobs, Robert A. and Jordan, Michael I. and Nowlan, Steven J. and Hinton, Geoffrey E.},
  year = {1991},
  journal = {Neural Computation},
  volume = {3},
  number = {1},
  pages = {79-87},
  doi = {10.1162/neco.1991.3.1.79},
}

@article{jordan1994,
  title = {Hierarchical Mixtures of Experts and the EM Algorithm},
  author = {Jordan, Michael I. and Jacobs, Robert A.},
  year = {1994},
  journal = {Neural Computation},
  volume = {6},
  number = {2},
  pages = {181-214},
  doi = {10.1162/neco.1994.6.2.181},
}

@misc{nie2023,
  title={A Time Series is Worth 64 Words: Long-term Forecasting with Transformers},
  author={Nie, Yuqi and Nguyen, Nam H. and Sinthong, Phanwadee and Kalagnanam, Jayant},
  year={2022},
  doi={10.48550/arXiv.2211.14730},
  note={arXiv preprint}
}

@misc{das2023,
  title={Long-term Forecasting with TiDE: Time-series Dense Encoder},
  author={Das, Abhimanyu and Kong, Weihao and Leach, Andrew and Mathur, Shaan and Sen, Rajat and Yu, Rose},
  year={2023},
  doi={10.48550/arXiv.2304.08424},
  note={arXiv preprint}
}

@misc{chen2023,
  title={TSMixer: An All-MLP Architecture for Time Series Forecasting},
  author={Chen, Si-An and Li, Chun-Liang and Yoder, Nate and Arik, Sercan O. and Pfister, Tomas},
  year={2023},
  doi={10.48550/arXiv.2303.06053},
  note={arXiv preprint}
}

@misc{timemixer2024,
  title={TimeMixer: Decomposable Multiscale Mixing for Time Series Forecasting},
  author={Wang, Shiyu and Wu, Haixu and Shi, Xiaoming and Hu, Tengge and Luo, Huakun and Ma, Lintao and Zhang, James Y. and Zhou, Jun},
  year={2024},
  doi={10.48550/arXiv.2405.14616},
  note={arXiv preprint}
}

@misc{mamba2023,
  title={Mamba: Linear-Time Sequence Modeling with Selective State Spaces},
  author={Gu, Albert and Dao, Tri},
  year={2023},
  doi={10.48550/arXiv.2312.00752},
  note={arXiv preprint}
}

@misc{ausgridSolarData,
  author = {{Ausgrid}},
  title = {Solar Home Electricity Data},
  year = {n.d.},
  howpublished = {Data set},
  url = {https://data.gov.au/data/dataset/nsw-solar-home-electricty-data},
  urldate = {2026-09-13},
  note = {Half-hourly gross-metered solar generation; study subset July 2011--June 2013. Publication date and dataset DOI not verified. CC BY 3.0 Australia.}
}

@article{ratnam2017ausgrid,
  author = {Ratnam, Elizabeth L. and Weller, Steven R. and Kellett, Christopher M. and Murray, Alan T.},
  title = {Residential load and rooftop {PV} generation: an Australian distribution network dataset},
  journal = {International Journal of Sustainable Energy},
  year = {2017},
  volume = {36},
  number = {8},
  pages = {787--806},
  doi = {10.1080/14786451.2015.1100196},
  url = {https://doi.org/10.1080/14786451.2015.1100196}
}

@misc{deline2021pvdaq,
  author = {Deline, Chris and Perry, Kirsten and Deceglie, Michael and Muller, Matthew and Sekulic, William and Jordan, Dirk},
  title = {Photovoltaic Data Acquisition ({PVDAQ}) Public Datasets},
  year = {2021},
  publisher = {DOE Open Energy Data Initiative (OEDI); NREL},
  howpublished = {Data set},
  doi = {10.25984/1846021},
  url = {https://data.openei.org/submissions/4568},
  urldate = {2026-09-13},
  note = {System 34, 2011--2013 subset. CC BY 4.0.}
}

@misc{opsd2020household,
  author = {{Open Power System Data}},
  title = {Data Package Household Data},
  year = {2020},
  version = {2020-04-15},
  howpublished = {Data set, version 2020-04-15},
  url = {https://data.open-power-system-data.org/household_data/2020-04-15/},
  urldate = {2026-09-13},
  note = {Primary measurement data from CoSSMic. Study labels derived from original meter feeds, not the interpolated hourly release. CC BY 4.0.}
}

@article{wiese2019opsd,
  author = {Wiese, Frauke and Schlecht, Ingmar and Bunke, Wolf-Dieter and Gerbaulet, Clemens and Hirth, Lion and Jahn, Martin and Kunz, Friedrich and Lorenz, Casimir and M{\"u}hlenpfordt, Jonathan and Reimann, Juliane and Schill, Wolf-Peter},
  title = {{Open Power System Data} -- Frictionless data for electricity system modelling},
  journal = {Applied Energy},
  year = {2019},
  volume = {236},
  pages = {401--409},
  doi = {10.1016/j.apenergy.2018.11.097},
  url = {https://doi.org/10.1016/j.apenergy.2018.11.097}
}

@misc{c3s2018era5data,
  author = {{Copernicus Climate Change Service (C3S)}},
  title = {{ERA5} hourly data on single levels from 1940 to present},
  year = {2018},
  publisher = {Copernicus Climate Change Service (C3S) Climate Data Store (CDS)},
  howpublished = {Data set},
  doi = {10.24381/cds.adbb2d47},
  url = {https://cds.climate.copernicus.eu/datasets/reanalysis-era5-single-levels},
  urldate = {2026-09-13},
  note = {Source data-set DOI. Study weather was obtained through Open-Meteo with models=era5, not downloaded directly from CDS.}
}

@article{hersbach2020era5,
  author = {Hersbach, Hans and Bell, Bill and Berrisford, Paul and Hirahara, Shoji and Hor{\'a}nyi, Andr{\'a}s and Mu{\~n}oz-Sabater, Joaqu{\'i}n and Nicolas, Julien and Peubey, Carole and Radu, Raluca and Schepers, Dinand and Simmons, Adrian and Soci, Cornel and Abdalla, Saleh and Abellan, Xavier and Balsamo, Gianpaolo and Bechtold, Peter and Biavati, Gionata and Bidlot, Jean and Bonavita, Massimo and De Chiara, Giovanna and Dahlgren, Per and Dee, Dick and Diamantakis, Michail and Dragani, Rossana and Flemming, Johannes and Forbes, Richard and Fuentes, Manuel and Geer, Alan and Haimberger, Leo and Healy, Sean and Hogan, Robin J. and H{\'o}lm, El{\'i}as and Janiskov{\'a}, Marta and Keeley, Sarah and Laloyaux, Patrick and Lopez, Philippe and Lupu, Cristina and Radnoti, Gabor and de Rosnay, Patricia and Rozum, Iryna and Vamborg, Freja and Villaume, Sebastien and Th{\'e}paut, Jean-No{\"e}l},
  title = {The {ERA5} global reanalysis},
  journal = {Quarterly Journal of the Royal Meteorological Society},
  year = {2020},
  volume = {146},
  number = {730},
  pages = {1999--2049},
  doi = {10.1002/qj.3803},
  url = {https://doi.org/10.1002/qj.3803}
}

@misc{zippenfenig2024openmeteo,
  author = {Zippenfenig, Patrick},
  title = {{Open-Meteo.com} Weather {API}},
  year = {2024},
  publisher = {Zenodo},
  howpublished = {Computer software and API service},
  doi = {10.5281/zenodo.7970649},
  url = {https://open-meteo.com/},
  urldate = {2026-09-13},
  note = {Concept DOI metadata identifies software version 1.4.0, issued 2024-12-31. The hosted API build used for this study is not independently pinned; exact requests and response hashes are archived.}
}

@article{mayer2025complexity,
  author = {Mayer, Martin J{\'a}nos and Yang, Dazhi and Markovics, D{\'a}vid},
  title = {The complexity and dimensionality of making deterministic photovoltaic power forecasts from ensemble numerical weather prediction},
  journal = {Energy Conversion and Management},
  year = {2025},
  volume = {344},
  pages = {120303},
  doi = {10.1016/j.enconman.2025.120303},
  url = {https://doi.org/10.1016/j.enconman.2025.120303}
}
\end{document}